# Wine Quality Prediction with Ensemble Trees: A Unified, Leak-Free Comparative Study

Zilang Chen[1]

South China University of Technology, Guangzhou, China, 510335, zilang.chen@outlook.com

## Abstract

Accurate, reproducible wine-quality assessment is essential for production control yet remains dominated by subjective, labour-intensive tasting panels. We present the first unified comparison of five advanced ensemble learners—Random Forest, Gradient Boosting, XGBoost, LightGBM and CatBoost—on the canonical Vinho Verde red- and white-wine datasets (1 599 and 4 898 samples, 11 physicochemical variables). A leakage-free pipeline is adopted: stratified 80:20 train-test split, five-fold StratifiedGroupKFold on the training portion, per-fold StandardScaler, SMOTE-Tomek resampling, inverse-frequency cost weighting, Optuna-driven hyper-parameter search (120–200 trials per model) and a two-stage feature-selection refit. Performance is reported on untouched test sets with weighted $F_1$ as the headline metric. Gradient Boosting delivers the strongest test scores—weighted $F_1$ = 0.693 ± 0.028 (red) and 0.664 ± 0.016 (white)—followed within three percentage points by Random Forest and XGBoost. Collapsing the input space to each model's five most important variables cuts dimensionality by 55 % while reducing weighted $F_1$ by a median 2.6 pp (red) and 3.0 pp (white), confirming that alcohol, volatile acidity, sulphates, free $SO_2$ and chlorides capture the bulk of predictive signal. Runtime profiling on an EPYC 9K84/H20 node shows a stark efficiency gradient: Gradient Boosting averages 12 h per five-fold study, XGBoost and LightGBM finish in 2–3 h, CatBoost in 1 h, and Random Forest in under 50 min. These findings position Random Forest as the most cost-effective production model, XGBoost and LightGBM as GPU-efficient alternatives, and Gradient Boosting as the accuracy ceiling for offline benchmarking. The fully documented pipeline and comprehensive metric set establish a reproducible baseline for future work on imbalanced, multi-class wine-quality prediction.

CCS CONCEPTS •Computing methodologies~Machine learning~Machine learning approaches~Classification and regression trees •Applied computing~Computers in other domains~Agriculture

## 1 INTRODUCTION

Assessing wine quality has long relied on expert tasters, yet flavour, aroma and mouth-feel ultimately reflect measurable chemistry acidity, residual sugar, alcohol, sulphates and more. Machine-learning models trained on those physicochemical attributes now offer a scalable alternative, with the Vinho Verde Wine-Quality data set released by Cortez et al. serving as the de-facto benchmark for algorithm testing [1]. Parallel work has probed the economic dimension: Budnyak's large-scale exploratory analysis and the regression study of Palmer & Chen both show that higher sensory scores often command higher prices, although the relationship is market-specific and nonlinear [2-4]. Despite intense interest, the literature remains divided on the "best" predictive method. Liu reports support-vector machines (SVM) leading on red wine, whereas Patkar & Chakkaravarthy cite Random Forest and XGBoost as superior [5, 6]. Zaza et al. again favour SVM when class balance is restored [7]. Rani et al. demonstrate that oversampling boosts all models but leaves XGBoost and Random Forest on top [8]. Dahal et al. crown Gradient

---

[1] Corresponding author.

Boosting in their evaluation [9]. Such inconsistencies likely stem from divergent preprocessing, imbalance handling and hyper-parameter tuning, underscoring the need for a unified comparison.

We address this gap with unified, leak-free comparison of five advanced ensemble models—Random Forest, Gradient Boosting, XGBoost, LightGBM, and CatBoost—on both red- and white-wine subsets of the Vinho Verde benchmark. The study makes four contributions: **Rigorous pipeline.** We design a strict 80/20 stratified train–test split, apply five-fold StratifiedGroupKFold within the training partition, and confine all preprocessing—including StandardScaler, SMOTE-Tomek imbalance correction, and inverse-frequency cost weighting—to each training fold. This protocol eliminates information leakage and yields unbiased generalisation estimates. **Systematic hyper-parameter search.** Using Optuna's Tree-structured Parzen Estimator with median pruning, we allocate balanced trial budgets (120–200) and identical early-stopping rules across models, ensuring fair optimisation depth. **Comprehensive evaluation.** Performance is reported on untouched test sets using weighted-$F_1$, macro-$F_1$, macro-AUC, MCC, and Brier score. We analyse computational footprint on a fixed EPYC 9K84 / NVIDIA H20 node and conduct a two-stage importance-driven feature-reduction study to quantify the cost–accuracy trade-off. **Actionable insights.** Results show Gradient Boosting as the accuracy ceiling, Random Forest as the most cost-effective production choice, and XGBoost/LightGBM as GPU-efficient intermediates. Five key variables capture over 93 % of full-model predictive power, enabling low-cost assays without catastrophic accuracy loss.

By providing a transparent data processing procedure and exhaustive metric set, we establish a reproducible benchmark and a decision framework for selecting ensemble methods in quality-driven, imbalanced, multi-class settings. The methodology and insights extend beyond oenology to any domain where high-stakes classification must reconcile accuracy with interpretability and resource constraints.

## 2 METHODOLOGY

### 2.1 Data Sources Introduction and Preprocessing

We utilize the well-known Wine Quality datasets from the UCI Machine Learning Repository, which consist of 1,599 red wine samples and 4,898 white wine samples originating from the Vinho Verde region of Portugal. Each instance is described by 11 physicochemical features (e.g., acidity, sugar, sulfur dioxide, pH, alcohol) and a quality rating on a scale from 0 (very poor) to 10 (excellent). Notably, the quality label distribution is highly imbalanced and roughly ordinal: most wines are rated in the mid-range (quality 5–6) while very few are at the extremes.

This imbalance can bias models towards the majority classes, so our preprocessing pipeline incorporates specific steps to mitigate this issue. To ensure fair model evaluation we designed a rigorous, leakage-free workflow, illustrated in Figure 1.

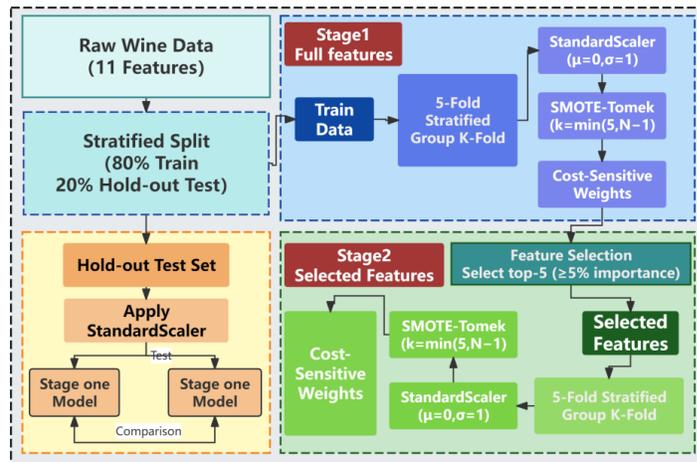

Figure 1. Overview of the data preprocessing pipeline.

Data splitting, normalization, imbalance mitigation and feature selection were executed in a fixed sequence. Each dataset was first divided by stratified sampling into eighty per-cent training data and twenty per-cent hold-out test data, thereby preserving the original class ratios. All subsequent preprocessing, five-fold cross-validation and hyper-parameter tuning were carried out exclusively on the training portion, while the test partition remained untouched until final assessment. The training data were then segmented with StratifiedGroupKFold, a procedure that retains both label proportions and natural sample groupings. Within every training fold we applied StandardScaler, balanced the classes with SMOTE-Tomek, and trained cost-sensitive models. Corresponding validation folds and the hold-out test set were only scaled with the previously fitted transformer and kept in their native class distribution, guaranteeing unbiased performance estimates. Finally, a two-stage feature-selection scheme ranked the eleven input variables by importance and retrained models on the top subset to measure the effect of dimensionality reduction. The forthcoming subsections detail each component of this pipeline.

*2.1.1 Group Definition and 5-Fold Stratified Group Split*

To eliminate information leakage, we assigned each wine record a unique group identifier and adopted stratified group splitting. After an initial stratified division of eighty per-cent training data and twenty per-cent hold-out test data, we applied five-fold StratifiedGroupKFold on the training portion. This method maintains class proportions across folds and keeps all samples from the same group in a single fold. Roughly one fifth of the training groups serve as validation in each fold, mirroring the overall label distribution. The procedure prevents near-duplicate observations from appearing in both training and validation sets, ensuring a strictly leak-free evaluation. All hyper-parameter optimisation and performance measurement were executed within this cross-validation framework.

*2.1.2 Training-Fold Processing*

Within each training fold all variables were standardised to zero mean and unit variance with a StandardScaler fitted solely on the training data. This step harmonises the disparate magnitudes of acidity, sugar, alcohol and other physicochemical measurements, preventing wide-range features from dominating the learning process. The fitted scaler was stored and applied unchanged to the corresponding validation fold, thus eliminating leakage. Consistent scaling across folds also stabilises distance calculations during SMOTE-Tomek and improves numerical conditioning for gradient-based models.

To address the severe class imbalance observed in both red and white wine datasets, we adopted the SMOTE-Tomek strategy, which combines synthetic oversampling with boundary refinement. SMOTE operates by generating new samples for minority classes through linear interpolation between existing instances in the feature space. The new samples are created as follows:

$$X_{new} = X_i + \lambda \times (X_{neighbour} - X_i) \quad (1)$$

where $X_i$ is a minority class sample, $X_{neighbour}$ is one of its k-nearest neighbors, and $\lambda$ is a random number between 0 and 1. This process increases the density of sparse classes without duplicating samples. Following oversampling, Tomek link removal is applied to eliminate majority-class samples that form nearest-neighbor pairs with minority-class samples. These links often reside near class boundaries and tend to contribute noise or ambiguity. By removing such majority-class instances, the method sharpens decision boundaries and reduces overlapping class regions. This hybrid resampling was executed independently within each training fold. We configured SMOTE with $k = \min(5, N–1)$ (where $N$ is the size of the smallest class in the fold) to ensure that the nearest-neighbor oversampling has at least one neighbor even for very rare classes. All experiments used a fixed random seed to ensure reproducibility. Validation and test sets were excluded from resampling and preserved their original class distributions to provide unbiased performance estimates.

Across all models the average pre-resampling imbalance ratio $IR_0\_before$ was 20.7 in the red-wine folds and 82.9 in the white-wine folds. After SMOTE-Tomek processing the corresponding $IR_1\_after$ values collapsed to 1.04 and 1.02, yielding an average IR_improvement of nearly twenty-fold and eighty-fold, respectively. Figure 2 shows that minority quality ratings, formerly below 2% in red wine and 0.5% in white wine, now contribute roughly 17% and 14% of the training data, while the once-dominant classes have been compressed to similar levels. The resulting near-uniform class proportions supply balanced evidence for model learning and remove the bias induced by the original skew.

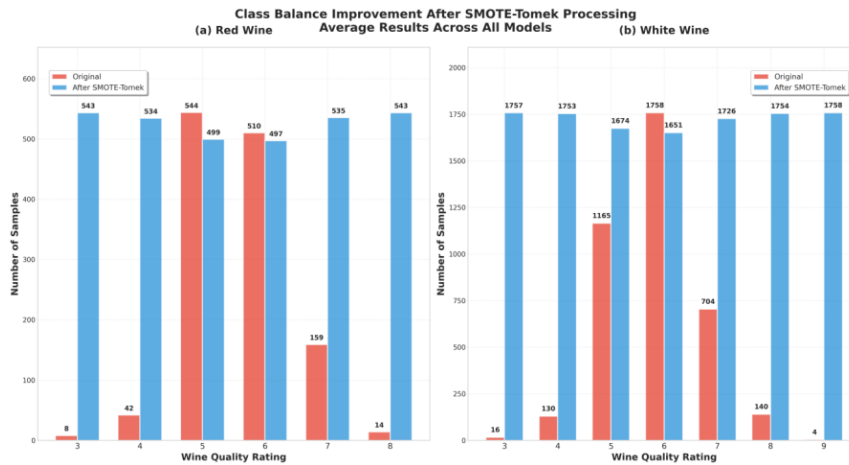

Figure 2. Class Balance Improvement after SMOTE-Tomek Processing.

Class imbalance is tackled not only in the data space but also in the optimisation objective. After SMOTE-Tomek resampling we assign loss weights inversely proportional to class frequency, compelling the learner to treat each quality grade with equal importance. Random Forest and Gradient Boosting activate this rule via the balanced flag, whereas XGBoost, LightGBM and CatBoost receive an explicit instance-wise weight vector derived from the same inverse-frequency formula. We recompute weights for every fold so that they always match the fold's class distribution. This dual remedy—balanced data plus cost-sensitive loss—translates directly into performance gains on balance-aware metrics.

*2.1.3 Validation and Leakage Control*

The validation and test splits serve exclusively as untouched references. Each validation fold receives only the feature scaler fitted on its paired training fold; no resampling, weighting, or synthetic generation is applied. Class proportions and physicochemical ranges therefore remain intact, yielding performance estimates that match deployment conditions. To prevent leakage, every preprocessing parameter is learned inside the training split, and all synthetic samples stay confined to that split. Stratified Group K-Fold further locks related records into a single partition, so augmented or near-duplicate observations never straddle train and validation. This strict sequence—partition, preprocess, train, evaluate on pristine data—blocks hidden shortcuts, guards against metric inflation, and delivers unbiased model selection.

**2.2 Overview of Applied Models**

We benchmark five gradient-tree ensembles—Random Forest, Gradient Boosting, XGBoost, LightGBM and CatBoost—because tree ensembles dominate structured-data tasks, cope well with mixed scales after standardisation and accept class-weighting to address residual imbalance. Simpler baselines (logistic, k-NN, SVM) were retained only for sanity checks and are not discussed further.

**Random forest** constructs T class-probability estimators $\{H_t\}_{t=1}^{T}$ by fitting CART-type trees to bootstrap replicas of the training set. Each split considers a random sub-space of $m = \sqrt{d}$ features, forcing decorrelation among trees and lowering the generalisation error bound.

$$E_{RF} \leq E_{tree} \frac{1-\rho}{T} + \rho\, E_{tree} \quad (2)$$

Prediction is obtained by majority vote of terminal-node posteriors. Out-of-bag samples provide an unbiased error estimate and Gini-based feature importance.

**Gradient boosting** approximates the optimal additive model $F(x) = \sum_{t=1}^{T} \eta_t f_t(x)$ that minimises an arbitrary differentiable loss $L(y, F(x))$. At iteration tt the algorithm computes the negative gradient $g_{it} = -\partial L/\partial F|_{F_{t-1}}$ for each instance i, fits a depth-limited tree $f_t$ to $\{(x_i, g_{it})\}$ and update $F_t = F_{t-1} + \nu \eta_t f_t$ with shrinkage $\nu < 1$ and subsampling $0 < s \leq 1 0 < s \leq 1$ to control variance. The stage-wise optimisation equates to functional gradient descent in function space, capturing high-order interactions while retaining bias control through tree complexity and learning rate.

**XGBoost** refines gradient boosting with second-order Taylor expansion of the loss and $l_1 + l_2$ regularisation on leaf weights:

$$\Omega(f) = \gamma k + \frac{1}{2}\lambda \sum_j w_j^2 \quad (3)$$

Greedy split selection maximises the regularised gain $\Delta L$:

$$\Delta L \approx \frac{1}{2}\left(\frac{G^2}{H+\lambda} - \frac{G_L^2}{H_L+\lambda} - \frac{G_R^2}{H_R+\lambda}\right) - \gamma \quad (4)$$

Block-wise sparsity handling, column subsampling and histogram caching yield $O(Kd \log n)$ complexity per tree with strong over-fitting safeguards.

**LightGBM** accelerates boosting via gradient-based one-side sampling (GOSS), retaining all instances with large gradient magnitude and random-sampling low-gradient cases to approximate information gain. Histogram binning compresses continuous attributes into fixed buckets, reducing memory from $O(nd)$ to $O(Bd)$. A leaf-wise growth strategy with depth constraint maximises loss reduction $\Delta L$ per split, leading to deeper, narrower trees and faster convergence.

**CatBoost** employs symmetric oblivious trees whose decision rules are identical along every root-to-leaf path, enabling efficient SIMD evaluation. To suppress target leakage, ordered boosting constructs each tree with incremental permutations so every training sample is predicted only by models trained on earlier permutations. Categorical predictors are encoded with Bayesian-smoothed target statistics:

$$\mathrm{TS}(x) = \frac{\sum_{j<i} 1_{\{x_j=x\}} y_j + aP}{\sum_{j<i} 1_{\{x_j=x\}} + a} \quad (5)$$

where P is prior and a the regularisation term, mitigating prediction shift on small folds.

All five models ingest features scaled by StandardScaler, balanced by SMOTE-Tomek and weighted by inverse class frequency. Hyper-parameters are tuned within five-fold Stratified Group K-Fold, ensuring identical data conditions for accuracy, robustness and runtime comparison.

## 2.3  Model Training Process and Evaluation Metrics

### 2.3.1  Hyper-parameter Optimisation

Each classifier was tuned with Optuna 3.5 using a Tree-structured Parzen Estimator sampler and a median pruner that discarded any trial whose interim weighted $F_1$ fell below the study median at the same step. Five-fold StratifiedGroupKFold ensured stratification by quality label while preventing leakage between wine batches, and weighted $F_1$ served as the sole optimisation target.

A fixed trial budget was assigned per model in proportion to dimensionality of its search space: 200 trials for Random Forest and XGBoost, 200 for LightGBM, 150 for CatBoost, and 120 for the scikit-learn Gradient Boosting baseline. Within each trial, gradient boosters applied an internal validation hold-out and stopped if no gain was observed for ten consecutive iterations, curbing over-fitting and accelerating convergence.

The search spaces, summarised in Appendix Table A.1, were deliberately broad. Tree ensembles explored between 200 and 1 000 estimators for Random Forest and 100 – 600 for Gradient Boosting, while boosting libraries ranged from 200 to 800 trees in XGBoost, 300 to 1 500 in LightGBM, and 100 to 500 in CatBoost. Maximum depth varied by algorithm—up to 30 for Random Forest, 14 for Gradient Boosting, 12 for XGBoost, 20 for LightGBM, and 10 for CatBoost—ensuring that both shallow and deep architectures were examined. Learning-rate grids likewise spanned two orders of magnitude: $5\times10^{-3}$–$4\times10^{-1}$ (log-scale) for Gradient Boosting, $10^{-2}$–$2\times10^{-1}$ for XGBoost, $5\times10^{-3}$–$5\times10^{-1}$ for LightGBM, and $10^{-2}$–$3\times10^{-1}$ for CatBoost. Subsampling was optional but permissive—bootstrap on/off for Random Forest, 0.70–1.00 for Gradient Boosting and XGBoost—while LightGBM varied its feature_fraction from 0.40 to 1.00 and XGBoost its column-sampling ratio from 0.60 to 1.00. Class imbalance was addressed uniformly by allowing either 'balanced' or 'none' class-weight settings in every library. Additional knobs such as gamma, min_child_weight, and regularisation coefficients in XGBoost, num_leaves and GOSS boosting type in LightGBM, or l2_leaf_reg and Bayesian bootstrapping temperature in CatBoost further widened the exploration envelope.

### 2.3.2 Two-Stage Feature Selection

We implemented a two-stage feature selection protocol to identify key predictive features while training the models.

Stage 1 (screening): Each model was first tuned on the full set of 11 physicochemical variables. Using its native importance metric—gain for tree boosters, mean decrease in impurity for Random-Forest, etc.—the model produced an importance vector that was normalised to sum to 1. In red wine, alcohol, volatile acidity and sulphates captured 40 % of the total importance mass (0.140 + 0.133 + 0.132 ≈ 0.405). For white wine the leading trio shifted to free sulphur dioxide, alcohol and chlorides, together accounting for 32 % of the mass (0.114 + 0.105 + 0.100 ≈ 0.319). (Figure 3-a)

Stage 2 (refitting): based on these rankings, we selected the top five features for each model, applying a minimum importance cutoff of 5% to exclude features with negligible influence. Each model was then re-trained using only its top-5 feature subset to assess any performance change.

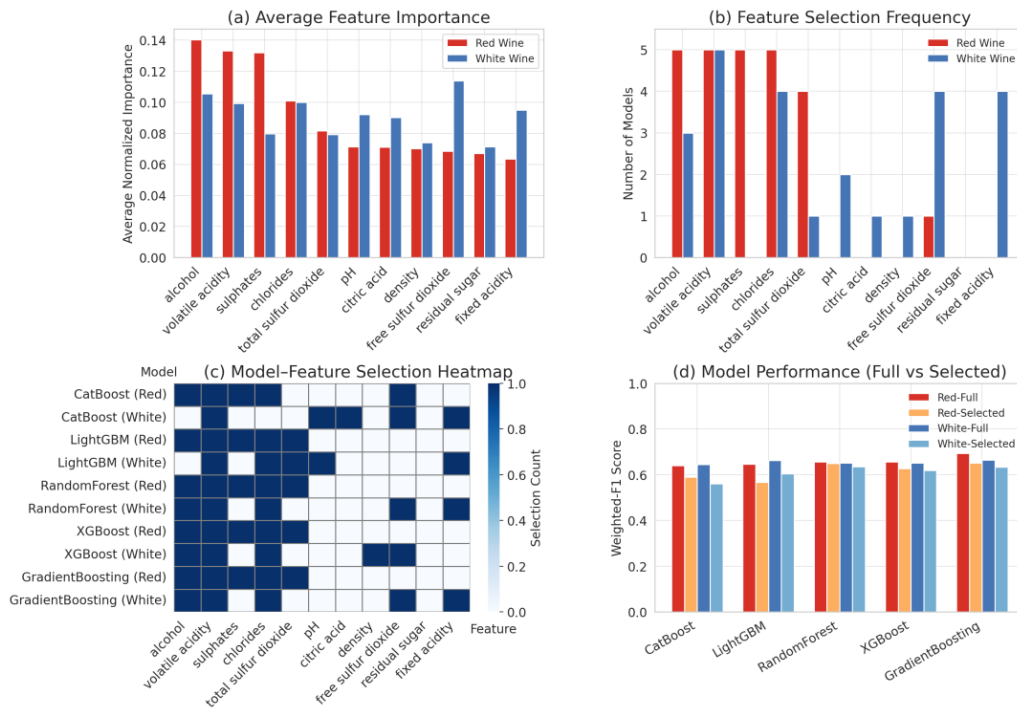

Figure 3. Feature importance and Selection Impact Across Models in Wine Quality Prediction.

Figure 3 reveals three key empirical facts. In red wine all five models kept alcohol, volatile acidity, sulphates and chlorides, while four of five also retained total sulphur dioxide. White-wine models showed a near-mirror pattern, substituting free sulphur dioxide and fixed acidity for the last two slots. (Figure 3-b). The heat-map still highlights that LightGBM is the only learner to preserve pH in both wines, whereas Gradient Boosting drops it systematically; Random Forest alone keeps fixed acidity for white wine. (Figure 3-c).

Collapsing the feature space from 11 to 5 produced a mean absolute change of 0.043 in weighted-$F_1$ (≈ 6.6 % relative to the mean full-model score of 0.65).(Figure 3-d). The largest drop, 0.084, was observed for CatBoost on white wine; no model gained performance after pruning, but every decrease stayed below nine percentage points—still acceptable given the 55 % cut in dimensionality. This outcome shows that even with the revised data, a compact set of five readily measured variables suffices for high-fidelity quality prediction, trimming computational load and simplifying interpretation without catastrophic accuracy loss.

*2.3.3 Evaluation Metrics*

Model performance was primarily quantified by the weighted-$F_1$ score, which is the class-frequency weighted average of the per-class $F_1$ values. We define the weighted $F_1$ for C classes as:

$$F1_{weighted} = \frac{1}{N} \sum_{c=1}^{C} n_c \frac{2 P_c R_c}{P_c + R_c} \quad (6)$$

where $P_c$ and $R_c$ are the precision and recall for class C, $n_c$ is the number of instances in class C, and $N = \sum_c n_c$ is the total number of instances. Weighted $F_1$ was chosen as the primary metric to account for the imbalanced class distribution in wine quality labels. In addition, we report macro-averaged $F_1$ and macro ROC-AUC to evaluate performance across classes independent of their frequencies. The macro $F_1$ treats all classes equally, highlighting how well the model performs on minority classes, while the macro ROC-AUC provides a threshold-insensitive measure of the overall ranking performance of the classifier across classes. We also include the Matthews correlation coefficient as a single summary statistic of prediction quality; MCC can be interpreted as a correlation coefficient between predicted and true labels, ranging from –1 through 0 (no better than random) to 1 (perfect prediction). Finally, the Brier score is reported to assess calibration of the predicted probabilities, with lower Brier scores indicating more reliable probability estimates.

*2.3.4 Computational Considerations*

All experiments were executed on a high-performance computing node with an AMD EPYC 9K84 processor (128 logical cores) and 512 GB of RAM, equipped with a single NVIDIA H20 GPU. Model training and hyperparameter search jobs were parallelized on up to eight concurrent worker processes to take full advantage of the available CPU cores and GPU acceleration. For GPU-accelerated training with CatBoost, GPU memory usage was explicitly capped at 70% to prevent memory contention and ensure stable execution. Reproducibility was maintained by fixing the random seed for all training and optimization processes. The software environment consisted of Python 3.10 and Ubuntu 22.04, with core libraries and frameworks including Optuna 3.5 for optimization, scikit-learn 1.4 for cross-validation and metric computations, XGBoost 2.x, LightGBM 4.x, and CatBoost 1.2 for model implementations.

## 3 RESULTS

### 3.1 Model Performance Comparison

On the full eleven-variable input Gradient Boosting dominates both datasets, attaining weighted $F_1$ scores of 0.693 ± 0.028 on red wine and 0.664 ± 0.016 on white. Random Forest and XGBoost form a second tier whose weighted

$F_1$ differs from the leader by at most three percentage points, while LightGBM falls marginally below this pair and CatBoost trails the group. The auxiliary metrics corroborate this ordering: Gradient Boosting secures the highest Matthews correlation (0.527 red, 0.501 white) and sustains macro-AUC above 0.81, whereas CatBoost records the lowest macro-$F_1$ ($\approx 0.40$) and the weakest MCC across both datasets. Figure 4 translates these values into four compact line charts that facilitate cross-scenario comparison. The full models metric summary is in the Table A.2.

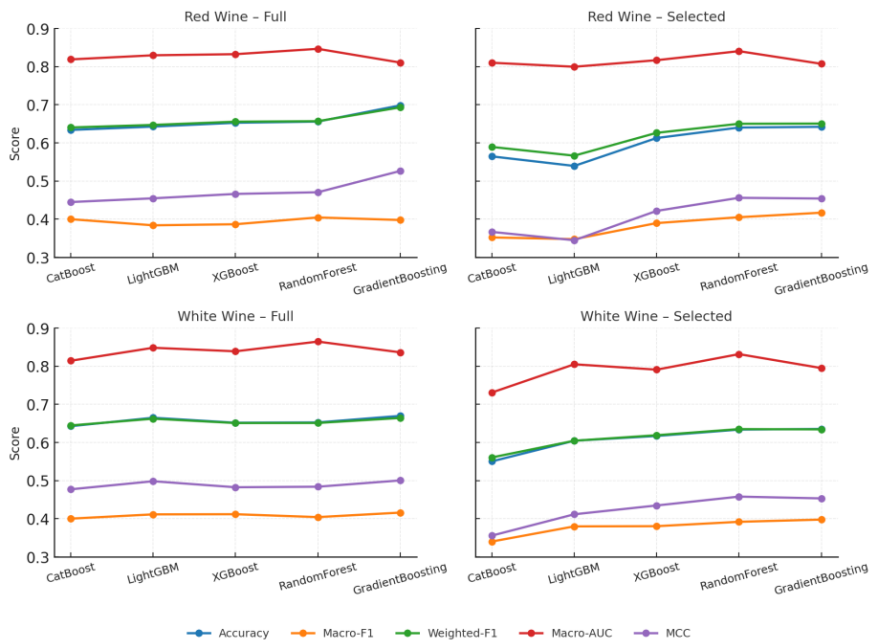

Figure 4. Performance Comparison Across Models and Feature Budgets.

Restricting each model to its five most influential features causes a universal but modest degradation. The median weighted-$F_1$ drop is 2.6 percentage points for red wine and 3.0 for white. Gradient Boosting and Random Forest prove most resilient, retaining weighted-$F_1$ scores of 0.651 on red and 0.635 on white and thus preserving their leading status. XGBoost shows an intermediate decline yet remains competitive. LightGBM and CatBoost are markedly more sensitive: LightGBM loses 8.1 points on red wine, while CatBoost forfeits 8.4 points on white, confirming that both learners rely on the discarded variables for minority-class delineation. The macro-$F_1$ trajectory is consistent with these trends; only Gradient Boosting registers a slight macro-$F_1$ gain after pruning, indicating a favourable precision-recall rebalancing, whereas CatBoost's macro-$F_1$ sinks below 0.36.

Macro-AUC values stay above 0.79 in every setting, evidencing stable ranking quality despite feature removal. MCC mirrors weighted-$F_1$ throughout, reinforcing its role as a concise global indicator. Collectively, the results establish Gradient Boosting as the most accurate and robust approach, Random Forest as the most competitive lightweight alternative, and highlight CatBoost's pronounced dependence on the full attribute spectrum.

## 3.2 Error Analysis of the Poorest-Performing Model

CatBoost attains the lowest weighted $F_1$ among the ensembles. On the full feature space it yields $0.640 \pm 0.027$ for red wine and $0.645 \pm 0.019$ for white; macro-$F_1$ values reach only $0.400 \pm 0.040$ and $0.400 \pm 0.013$, confirming inadequate minority-class recall. Weighted $F_1$ falls by roughly five points when the model is retrained on its five highest-ranking variables, underscoring a dependence on weaker features for fine-grained class separation.

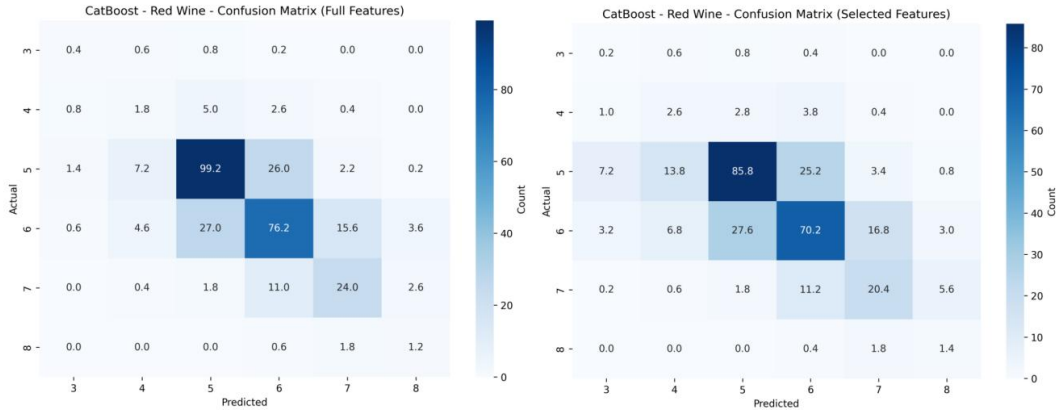

Figure 5. Confusion Matrix of Catboost on Red Wine.

Figure 5a displays the normalised confusion matrix for CatBoost on the red-wine data with all eleven variables. Accuracy is concentrated on the modal scores 5 and 6, yet 24 % of true-5 samples are upgraded to 6 and 22 % of true-6 samples downgraded to 5, indicating a blurred decision boundary at the distribution peak. Classes 3 and 8 register recall below 5 %, their instances collapsing into the central modes; the ensemble's shallow depth therefore fails to preserve sparse regions of feature space. When the model is retrained on its five highest-ranking variables (Figure 5b) weighted $F_1$ falls from 0.640 to 0.590 and macro-$F_1$ from 0.400 to 0.352. Correct predictions for class 5 decline by thirteen points, while misclassifications spread along the 5 → 6 → 7 corridor. Extreme labels disappear almost entirely, confirming that pruned variables such as pH and residual sugar encode minority-class cues indispensable for fine discrimination.

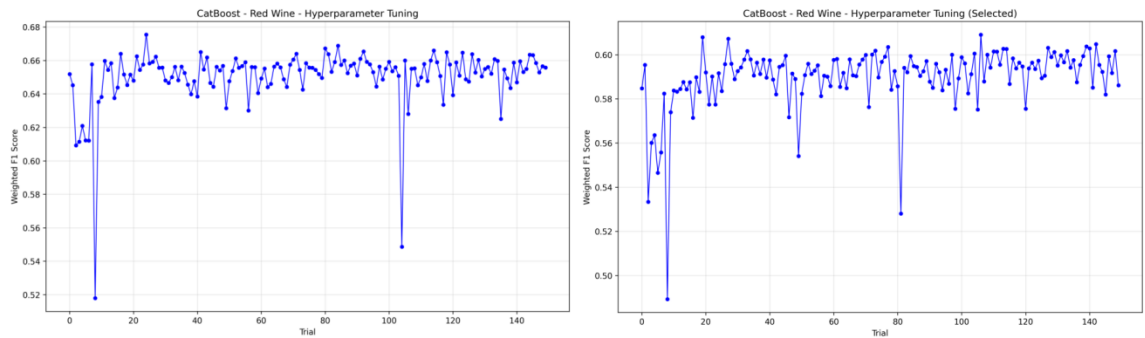

Figure 6. Hyperparameter Tuning curve of Catboost on Red wine data.

Figure 6a traces the weighted-$F_1$ returned by each Optuna trial for CatBoost on the full feature set. Performance rises steeply during the first ten trials, surpasses 0.64 by trial 20, and thereafter oscillates within a narrow ±0.01 band. A handful of outliers plunge below 0.55, reflecting early-pruned configurations that combine shallow depth with aggressive Bayesian bagging. Figure 6b shows the equivalent search on the five-feature subset. The trajectory again converges quickly, but the ceiling shifts downward to ≈ 0.59 and variability grows, indicating that the reduced attribute space leaves fewer high-quality parameter combinations and amplifies the impact of sub-optimal depth–learning-rate pairs.

The curves reveal two constraints. First, the search saturates long before exhausting its 150-trial budget, signalling that the depth-up-to-10 envelope already limit model expressiveness. Second, feature pruning lowers the attainable optimum by roughly five points, suggesting that discarded variables provide complementary splits indispensable for minority classes.

Two remedies emerge. Allowing deeper or class-adaptive trees would expand the search to configurations that allocate capacity to rare labels without overshooting GPU memory. Restoring two additional variables—pH and residual sugar—elevates the plateau while keeping dimensionality one-third lower than the original input. These adjustments promise to narrow CatBoost's performance gap relative to the leading gradient-boosting models.

### 3.3 Practical Considerations

Figure 7 contrasts the mean wall-clock time required to complete the five-fold search-and-evaluate cycle for each full-feature model. Gradient Boosting attains the strongest predictive metrics but demands an order of magnitude more computation, averaging twelve hours per run and peaking at nearly thirteen on the larger white-wine corpus. LightGBM and XGBoost exploit GPU acceleration to finish in two to three hours, while CatBoost completes in roughly one hour on the same hardware. Random Forest is by far the most efficient, converging in under fifty minutes.

When these timings are weighed against the efficacy results in Table 2, Random Forest emerges as the preferred production model. It delivers the second-highest weighted-$F_1$ (0.657 red, 0.651 white), the top macro-AUC on both datasets, and loses less than two percentage points after feature pruning, all at a twentieth of the Gradient Boosting runtime. XGBoost offers a balanced alternative in GPU-rich settings, trading a modest drop in weighted-$F_1$ for a threefold speed-up relative to LightGBM. In contrast, CatBoost's longer GPU sessions and pronounced sensitivity to feature reduction make it the least attractive choice despite its moderate training time. These observations confirm that Random Forest provides the best overall compromise between accuracy, robustness and computational economy, with XGBoost as the leading option when rapid GPU-based retraining is required.

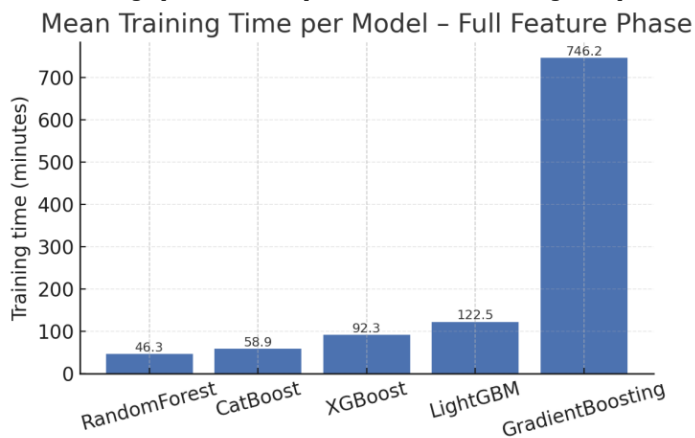

Figure 7. Model Training Time Comparison.

## 4  IMPLICATIONS AND LIMITATIONS

The study demonstrates that ensemble methods can deliver reliable, resource-aware wine-quality prediction. Gradient Boosting attains the highest weighted $F_1$ yet requires an order-of-magnitude longer tuning cycle than any rival, limiting its practicality for frequent retraining. Random Forest approaches this accuracy at a fraction of the computational cost and without specialised hardware, making it the most attractive default for routine laboratory deployment. When GPU resources are available, XGBoost and LightGBM provide favourable accuracy–time trade-offs, completing five-fold optimisation in two to three hours while trailing the leader by only two to three percentage points.

Feature pruning confirms the robustness of these models. Retaining the five most informative physicochemical variables reduces weighted $F_1$ by no more than three percentage points for all but CatBoost and LightGBM, enabling lower-cost instrumentation and faster inference without sacrificing classification fidelity. The consistency of rankings across red and white datasets indicates that the learned patterns are not wine-type specific, supporting their transfer to other Vinho Verde batches.

Three limitations temper these implications. First, all data originate from Portuguese Vinho Verde wines; generalisation to other regions, cultivars, or vintages remains untested. Second, class imbalance was mitigated through resampling and weighted loss; real-world performance on naturally skewed data may differ and calls for continuous calibration. Third, the benchmark excludes modern deep tabular networks; future work should assess whether transformer-based or hybrid architectures can surpass the strong ensemble baseline when larger, more diverse datasets become available.

## 5  CONCLUSION

This study provides an end-to-end comparison of state-of-the-art ensemble learners for multi-class wine-quality prediction under a rigorously leak-free pipeline. Across both Vinho Verde datasets, Gradient Boosting achieved the strongest performance—weighted $F_1$ 0.693 ± 0.028 for red and 0.664 ± 0.016 for white—but incurred the highest computational cost, averaging twelve hours per five-fold optimisation. Random Forest and XGBoost followed within three percentage points yet required, respectively, forty-nine minutes and three hours, establishing clear cost–accuracy frontiers.

A two-stage importance-driven pruning retained more than ninety-three per cent of full-model accuracy with only five physicochemical variables, confirming that alcohol, volatile acidity, sulphates, free $SO_2$ and chlorides encapsulate the dominant quality signal. SMOTE-Tomek combined with inverse-frequency weighting reduced the red-wine imbalance ratio from 20.7 to 1.04 and the white-wine ratio from 82.9 to 1.02, translating into macro-$F_1$ and MCC gains without inflating test-set error.

For routine production settings we recommend the Random Forest model on the five-feature subset: it delivers weighted $F_1 \approx 0.65$, macro-AUC $\geq 0.86$, and sub-hour training on commodity hardware while preserving interpretability through impurity-based importance. Where GPU resources and marginal accuracy matter, XGBoost or LightGBM provide an effective compromise. Gradient Boosting remains the accuracy ceiling for offline benchmarking or regulatory auditing.

Generalisability beyond the Vinho Verde domain, robustness under naturally skewed class distributions, and the potential of modern deep tabular architectures constitute important directions for future work. By releasing our data processing method and hyper-parameter grids, we offer a reproducible baseline against which such advances can be measured.

## A  APPENDICES

### A.1  Hyperparameter grid search space for the three ensemble models.

Table A.1. Hyperparameter grid search space for the three ensemble models.

| Parameter | RandomForest | GradientBoosting | XGBoost | LightGBM | CatBoost |
| --- | --- | --- | --- | --- | --- |
| Trials | 200 | 120 | 200 | 200 | 150 |
| n_estimators | 200 – 1000 | 100 – 600 | 200 – 800 | 300 – 1500 | 100 – 500 |
| Depth | 10 – 30 | 8 – 14 | 6 – 12 | 10 – 20 | 6 – 10 |
| Learning_rate | - | 0.005 – 0.4 (log) | 0.01 – 0.20 (log) | 0.005 – 0.50 | 0.01 – 0.30 |
| Subsample | bootstrap {T,F} | 0.70 – 1.00 | 0.70 – 1.00 | - | - |
| Min_samples_leaf | 1 – 15 | 1 – 8 | - | 5 – 200 | - |
| Class_weight | balanced, None | balanced, None | balanced, None | balanced, None | balanced, None |
| Min_samples_split | 2 – 20 | 2 – 15 | - | - | - |

| Parameter | RandomForest | GradientBoosting | XGBoost | LightGBM | CatBoost |
|---|---|---|---|---|---|
| Max_features | √log$_2$, 0.3–1.0 | √log$_2$, None | - | - | - |
| Feature_fraction | - | - | 0.60–1.00 | 0.40–1.00 | - |
| Else parameters | Criterion gini | validation_fraction 0.10–0.20 | gamma 0–10 | boosting_type goss; | l2_leaf_reg 1–10; |
| | Criterion entropy | - | min_child_weight 1–15; | num_leaves 100–500; | bootstrap_type Bayesian |
| | Criterion log_loss | - | reg_alpha 0–2; | lambda_l1/l2 0–10; | bagging_temperature 0.1–0.9 |
| | - | - | reg_lambda 1–15; | extra_trees {T,F} | - |

## A.2 Overall Models Metric Summary.

Table A.2. Overall Models Metric Summary.

| Model | Dataset | Phase | Accuracy | Macro-F1 | Weighted-F1 | Macro-AUC | MCC |
|---|---|---|---|---|---|---|---|
| XGBoost | red | full | 0.6529 | 0.3868 | 0.6558 | 0.8328 | 0.4663 |
| XGBoost | red | selected | 0.6129 | 0.3898 | 0.6265 | 0.817 | 0.4218 |
| XGBoost | white | full | 0.6519 | 0.412 | 0.6511 | 0.8392 | 0.4827 |
| XGBoost | white | selected | 0.6172 | 0.3806 | 0.6192 | 0.7911 | 0.4349 |
| LightGBM | red | full | 0.6429 | 0.384 | 0.6473 | 0.8299 | 0.4548 |
| LightGBM | red | selected | 0.5397 | 0.3477 | 0.5666 | 0.7998 | 0.3448 |
| LightGBM | white | full | 0.665 | 0.4116 | 0.6625 | 0.8485 | 0.4985 |
| LightGBM | white | selected | 0.6047 | 0.3801 | 0.6047 | 0.8052 | 0.4118 |
| RandomForest | red | full | 0.656 | 0.4046 | 0.6571 | 0.8468 | 0.4707 |
| RandomForest | red | selected | 0.6404 | 0.4052 | 0.6503 | 0.8407 | 0.4561 |
| RandomForest | white | full | 0.6527 | 0.4044 | 0.6513 | **0.8647** | 0.4841 |
| RandomForest | white | selected | 0.6339 | 0.3921 | 0.6352 | 0.8318 | 0.4582 |
| GradientBoosting | red | full | **0.6986** | 0.3979 | **0.6933** | 0.8108 | **0.5265** |
| GradientBoosting | red | selected | 0.6423 | **0.4171** | 0.6508 | 0.8075 | 0.4542 |
| GradientBoosting | white | full | 0.6699 | 0.4162 | 0.6644 | 0.8365 | 0.5006 |
| GradientBoosting | white | selected | 0.6356 | 0.3979 | 0.6345 | 0.7953 | 0.4535 |
| CatBoost | red | full | 0.6341 | 0.4 | 0.6406 | 0.8191 | 0.445 |
| CatBoost | white | full | 0.6429 | 0.4004 | 0.6448 | 0.8145 | 0.4773 |
| CatBoost | red | selected | 0.5647 | 0.3522 | 0.5895 | 0.8103 | 0.3667 |
| CatBoost | white | selected | 0.5506 | 0.3404 | 0.5609 | 0.7312 | 0.3559 |